\documentclass[
]{ceurart}

\usepackage{listings}
\usepackage[nolist]{acronym}

\newcommand{\todo}[1]{} 
\renewcommand{\todo}[1]{{\noindent \color{red} \textbf{TODO: } {#1}}}

\begin{document}

\copyrightyear{2025}
\copyrightclause{Copyright for this paper by its authors.
  Use permitted under Creative Commons License Attribution 4.0
  International (CC BY 4.0).}

\conference{Preprint under review for CLEF 2025 Working Notes, September 9 -- 12 September 2025, Madrid, Spain}

\title{INESC-ID @ eRisk 2025: Exploring Fine-Tuned, Similarity-Based, and Prompt-Based Approaches to Depression Symptom Identification}
\title[mode=sub]{Notebook for the eRisk Lab at CLEF 2025}


\author[1,2]{Diogo A.P. Nunes}[%
orcid=0000-0002-6614-8556,
email=diogo.p.nunes@inesc-id.pt,
]
\cormark[1]
\fnmark[1]

\author[1,3]{Eugénio Ribeiro}[%
orcid=0000-0001-7147-8675,
email=eugenio.ribeiro@inesc-id.pt,
]
\cormark[1]
\fnmark[1]

\address[1]{INESC-ID Lisboa, Rua Alves Redol 9, 1000-029 Lisboa, Portugal}
\address[2]{Instituto Superior Técnico, Universidade de Lisboa, Av. Rovisco Pais, 1049-001 Lisboa, Portugal}
\address[3]{Instituto Universitário de Lisboa (ISCTE-IUL), Avenida das Forças Armadas, 1649-026 Lisboa, Portugal}

\cortext[1]{Corresponding author.}
\fntext[1]{These authors contributed equally.}

\begin{acronym}
    \acro{AP}{Average Precision}
    
    \acro{BDI}{Beck's Depression Inventory - II}

    \acro{CNN}{Convolutional Neural Network}

    \acro{EMS}{Early Maladaptive Schemas}

    \acro{FCT}{Fundação para a Ciência e a Tecnologia}
    
    \acro{IR}{Information Retrieval}
    
    \acro{LIWC}{Linguistic Inquiry and Word Count}
    \acro{LLM}{Large Language Model}

    \acro{MAP}{Mean Average Precision}

    \acro{NLP}{Natural Language Processing}
    \acro{NDCG@1000}{Normalized Discounted Cumulative Gain @1000}

    \acro{P@10}{Precision @10}

    \acro{RSDD}{Reddit Self-reported Depression Diagnosis}
    \acro{R-PREC}{R-Precision}
    
    \acro{SMHD}{Self-reported Mental Health Diagnoses}

    \acro{WHO}{World Health Organization}
    
\end{acronym}
\begin{abstract}
In this work, we describe our team's approach to eRisk's 2025 Task 1: Search for Symptoms of Depression. Given a set of sentences and the \ac{BDI} questionnaire, participants were tasked with submitting up to $1,000$ sentences per depression symptom in the \ac{BDI}, sorted by relevance. Participant submissions were evaluated according to standard \ac{IR} metrics, including \ac{AP} and \ac{R-PREC}. The provided training data, however, consisted of sentences labeled as to whether a given sentence was relevant or not w.r.t. one of \ac{BDI}'s symptoms. Due to this labeling limitation, we framed our development as a binary classification task for each \ac{BDI} symptom, and evaluated accordingly. To that end, we split the available labeled data into training and validation sets, and explored foundation model fine-tuning, sentence similarity, \ac{LLM} prompting, and ensemble techniques. The validation results revealed that fine-tuning foundation models yielded the best performance, particularly when enhanced with synthetic data to mitigate class imbalance. We also observed that the optimal approach varied by symptom. Based on these insights, we devised five independent test runs, two of which used ensemble methods. These runs achieved the highest scores in the official \ac{IR} evaluation, outperforming submissions from $16$ other teams.
\end{abstract}


\maketitle

\acresetall

\section{Introduction}
\label{sec:introduction}

Mental health is central to the overall physical health. Indeed, other diseases are of increased risk in the presence of psychological disorders \cite{Prince2007}. Depression is one such disorder; it can be caused by both physiological and psychological factors, and its symptoms may include a depressive mood, lack of interest and pleasure, and reduced energy \cite{Paykel2008}. According to the \ac{WHO}\footnote{\url{https://www.who.int/news-room/fact-sheets/detail/depression}}, $5\%$ of the global population suffers from depression, with a higher incidence on women. Depression is also one of the most common comorbidities of chronic diseases, such as cancer and chronic pain, in part because of their psychosocial burden; in these cases, the depression diagnosis is an increased challenge due to the overlapping symptoms and confounding factors \cite{Gold2020}.

The relation between mental disorders and the linguistic expression has been increasingly explored \cite{Yates2017, Cohan2018, Bridianne2021, Yahya2023}. In fact, depression symptoms manifest in patients' language commonly as short and directive communication, limited development of concepts, self-focused attention, negative sentiment, verbosity of auxiliary terms, and disfluencies \cite{Bridianne2021, Yahya2023, Trifu2024}. This motivates the development of \ac{NLP} techniques to monitor and detect depression from language use. However, language is modulated by a plethora of factors beyond psychological and clinical states, namely demographic and sociocultural variables, which can be confounding factors towards that objective \cite{Trifu2024}.

Social media presents an opportunity for the development of monitoring and detection systems for depression in online platforms. These may allow for early detection and quick action on a large-scale, giving emergence to eRisk's task of sentence ranking for depression symptoms, which was introduced in 2023 \cite{Parapar2023}. Previous participant submissions to this and similar tasks included (key)word-based frequency features with downstream classification and ranking models \cite{Losada2017}, sentence embeddings for similarity ranking \cite{Recharla2023}, and, more recently, \acp{LLM} for synthetic dataset generation \cite{Ang2024}.


Our team's participation in this task comprised the exploration of multiple methods to select and rank relevant sentences for a given \ac{BDI} symptom. Although the official task evaluation was based on standard \ac{IR} metrics, we mainly framed our methods as binary classification or regression tasks due to training data limitations, as described below. Our methodology included the fine-tuning of foundation models, similarity-based ranking in an unsupervised setting, \ac{LLM} relevancy prompting and synthetic data generation, and ensemble techniques. We developed and validated these approaches in our training and validation splits of the provided labeled training dataset, based on classification metrics. 
A high-precision, ensemble run was our best performing submission in the official \ac{IR} evaluation, placed \textit{1st} among $17$ teams, for a total of $67$ runs. This paper describes our approach and its results in detail.
\section{Related Work}
\label{sec:related}

Focusing on text as an instantiation of language, previous work has attempted to identify the linguistic markers of depression. These are characteristics of language use that can be used to separate depression patients from controls. \citeauthor{Trifu2024} conducted such a study with $62$ patients diagnosed with major depressive disorder and $43$ controls. They sampled language use through prompted narratives on something that provided (or used to provide) pleasure. Participants' transcribed answers were analyzed with \ac{LIWC} \cite{Tausczik2009}, which is a proprietary knowledge- and dictionary-based psycholinguistic feature extractor. Their statistical analysis found that there were significant language use differences between patients and controls; for instance, depression patients used shorter sentences, and more frequently the personal pronoun plural (``we''), informal language, interrogations, and other punctuation in general. Their sentences were also more likely to be formed in the past tense. Semantically, depression patients were more likely to talk about biological processes, health, and money, and less about leisure. Other analyses observed similar findings \cite{Yahya2023}, laying the foundation for the type of information that should be monitored by systems for early detection of depression online.



eRisk's task \cite{Parapar2023, Parapar2024, Parapar2025} of ranking sentences for depression symptom detection in online platforms entails slightly different constraints from the related work above. It focuses on learning the relevancy of a given sentence for a given \ac{BDI} symptom. \ac{BDI} includes $21$ symptoms, such as sadness, pessimism, loss of pleasure, self-dislike, worthlessness, and agitation. For each symptom, four descriptions are provided, seemingly in order of intensity. Tab.~\ref{tab:bdi-example} shows two such examples. The two major constraints in this task are: 1) symptom-level detection is more granular than binary depression diagnosis, and 2) sentence-level detection of depression lacks in context w.r.t. user-level detection. Indeed, since the 2024 edition of this task, sentences were contextualized with previous and subsequent sentences; this, however, is still far from the context available for user-level detection of depression. Below, we briefly describe the approaches of the best performing teams in the past two editions.


\begin{table}
    \centering
    \caption{Examples of \ac{BDI} symptom options.}
    \label{tab:bdi-example}
    \begin{tabular}{@{} lll @{}}
         \toprule
         \textbf{Symptom} & Sadness & Crying \\
         \midrule
         \multirow{4}{*}{\textbf{Options}} & 0. I do not feel sad. & 0. I don't cry anymore than I used to. \\
         & 1. I feel sad much of the time. & 1. I cry more than I used to. \\
         & 2. I am sad all the time. & 2. I cry over every little thing. \\
         & 3. I am so sad or unhappy that I can't stand it. & 3. I feel like crying, but I can't. \\ 
         \bottomrule
    \end{tabular}
\end{table}

In 2023, \citeauthor{Recharla2023} submitted four runs, all unsupervised and similarity-based (notably, training data was not available in this edition, since it was the first). After pre-processing, they calculated two types of embeddings for each sentence and \ac{BDI} symptom option (locally trained Word2Vec \cite{Mikolov2013} and the pretrained paraphrase-MiniLM-L3-v2\footnote{\url{https://huggingface.co/sentence-transformers/paraphrase-MiniLM-L3-v2}} SentenceTransformer \cite{Reimers2019}). They selected and ranked the top $1,000$ most similar corpus sentences to each \ac{BDI} symptom, according to their average similarity to the symptom's options. They included both weighted and unweighted similarity averages (where the weight was given by the increasing intensity of the symptom's options; see Tab.~\ref{tab:bdi-example}). SentenceTransformer-based embeddings outperformed locally trained Word2Vec-based embeddings by a large margin. Overall, unweighted similarity average of SentenceTransformer-based embeddings performed the best.

\citeauthor{Ang2024} submitted five runs in 2024 \cite{Parapar2024}. All of their runs were also similarity-based. In order to calculate the relevance of a candidate sentence w.r.t. a \ac{BDI} symptom, they developed three sets of symptom exemplars. These included a set with the original \ac{BDI} symptom options (see Tab.~\ref{tab:bdi-example}), the previous set plus GPT-4 \cite{gpt4} synthesized exemplars based on \ac{EMS}, and the previous two sets plus synthetic exemplars demonstrating positive-sentiment user state (e.g., ``I'm sad'' \textit{versus} ``I'm happy'' for the Sadness symptom). They extracted embeddings of both candidate sentences and symptom exemplars with pretrained and fine-tuned SentenceTransformer models, including all-mpnet-base-v2\footnote{\url{https://huggingface.co/sentence-transformers/all-mpnet-base-v2}}, all-MiniLM-L12-v2\footnote{\url{https://huggingface.co/sentence-transformers/all-MiniLM-L12-v2}}, and all-distilroberta-v1\footnote{\url{https://huggingface.co/sentence-transformers/all-distilroberta-v1}}. Their fine-tuning was based on contrastive learning with annotated training data, which was officially available starting in 2024. Like \citeauthor{Recharla2023}, similarity was measured with average cosine-similarity. Although candidate sentence context was available in 2024, \citeauthor{Ang2024} do not report to have leveraged it in their approach. Their best performing run was an ensemble of various pretrained and fine-tuned sentence embeddings and symptom exemplars. Data from both previous editions and corresponding annotations were available for this year's edition.

         


         
\section{Methodology}
\label{sec:methodology}

In this section we describe in detail the experimental setup defining our approach to eRisk's ``Task 1: Search for Symptoms of Depression''. First, we discuss the official training and test data, and our own development training and validation splits. We then discuss our technical approach, encompassing foundation model fine-tuning, similarity-based methods, and \ac{LLM} prompting. Finally, we describe our regression/classification evaluation framework in contrast to the official evaluation based on \ac{IR} metrics.

\subsection{Dataset}

Participants were provided with official training and test splits of the dataset. All sentences in the dataset were presented in TREC format, and were characterized by a unique ID (\texttt{<DOCNO>}) and their text (\texttt{<TEXT>}). Some sentences were also characterized by their surrounding context, when available, i.e., the text of the previous and subsequent sentences (\texttt{<PRE>} and \texttt{<POST>}, respectively). 

The official training set comprised data from the two previous editions of this task \cite{Parapar2023, Parapar2024}. A portion of this set was labeled according to the task's annotation guidelines, i.e., whether a given sentence was relevant or not to a given \ac{BDI} symptom. In fact, two binary labels were provided per annotated sentence, one representing the annotators' majority vote, and the other the annotators' unanimous vote w.r.t. that relevancy. Not all annotated sentences were labeled for all \ac{BDI} symptoms. For development, we randomly split the official annotated subsection of the training set ($26,290$ sentences) in training (\textit{train}; $80\%$) and validation (\textit{val}; $20\%$) sets. We stratified the splits per symptom and per label (majority and unanimity). Tab.~\ref{tab:annotated-symp} shows the distribution of labels per \ac{BDI} symptom in our development splits. We purposefully mixed annotated sentences from the 2023 and 2024 editions in both \textit{train} and \textit{val} splits to, first, avoid annotation biases that might have occurred in any one of the previous editions, and second, improve data imbalances. 

\begin{table}
    \centering
    \caption{Number of annotated instances per \ac{BDI} symptom (number of instances with positive majority label / number of instances with positive unanimity label).}
    \label{tab:annotated-symp}
    \begin{tabular}{@{} llll @{}}
         \toprule
         Symptom & \multicolumn{1}{c}{\textit{train}} & \multicolumn{1}{c}{\textit{val}} & \multicolumn{1}{c}{Total} \\
         \midrule
Sadness & $1384$ ($541$/$303$) & $347$ ($136$/$76$) & $1731$ ($677$/$379$) \\
Pessimism & $1272$ ($457$/$169$) & $310$ ($113$/$44$) & $1582$ ($570$/$213$) \\
Past failure & $1015$ ($366$/$196$) & $258$ ($97$/$51$) & $1273$ ($463$/$247$) \\
Loss of pleasure & $1134$ ($302$/$157$) & $276$ ($71$/$36$) & $1410$ ($373$/$193$) \\
Guilty feelings & $875$ ($347$/$260$) & $211$ ($88$/$66$) & $1086$ ($435$/$326$) \\
Punishment feelings & $1026$ ($164$/$83$) & $248$ ($39$/$20$) & $1274$ ($203$/$103$) \\
Self-dislike & $963$ ($365$/$237$) & $221$ ($84$/$58$) & $1184$ ($449$/$295$) \\
Self-criticalness & $1015$ ($280$/$166$) & $256$ ($68$/$42$) & $1271$ ($348$/$208$) \\
Suicidal thoughts or wishes & $943$ ($471$/$357$) & $230$ ($116$/$86$) & $1173$ ($587$/$443$) \\
Crying & $886$ ($425$/$293$) & $222$ ($112$/$76$) & $1108$ ($537$/$369$) \\
Agitation & $1134$ ($408$/$248$) & $289$ ($101$/$60$) & $1423$ ($509$/$308$) \\
Loss of interest & $988$ ($225$/$114$) & $251$ ($57$/$32$) & $1239$ ($282$/$146$) \\
Indecisiveness & $1150$ ($300$/$149$) & $286$ ($74$/$38$) & $1436$ ($374$/$187$) \\
Worthlessness & $763$ ($188$/$140$) & $203$ ($47$/$35$) & $966$ ($235$/$175$) \\
Loss of energy & $936$ ($298$/$213$) & $241$ ($74$/$52$) & $1177$ ($372$/$265$) \\
Changes in sleeping pattern & $1088$ ($461$/$273$) & $280$ ($124$/$74$) & $1368$ ($585$/$347$) \\
Irritability & $902$ ($289$/$178$) & $226$ ($69$/$47$) & $1128$ ($358$/$225$) \\
Changes in appetite & $1011$ ($373$/$207$) & $244$ ($90$/$50$) & $1255$ ($463$/$257$) \\
Concentration difficulty & $794$ ($243$/$152$) & $207$ ($67$/$42$) & $1001$ ($310$/$194$) \\
Tiredness or fatigue & $781$ ($269$/$173$) & $189$ ($65$/$37$) & $970$ ($334$/$210$) \\
Loss of interest in sex & $979$ ($341$/$173$) & $256$ ($82$/$40$) & $1235$ ($423$/$213$) \\
        \midrule
\textbf{Total} & $21039$ ($7113$/$4241$) & $5251$ ($1774$/$1062$) & $26290$ ($8887$/$5303$) \\
         \bottomrule
    \end{tabular}
\end{table}

During the data exploration stage, we noticed duplicated sentences in the official training set, although with varying capitalization or formatting (e.g., ``I'm sad'' and ``i'm sad.''). These duplicates were not always coherently annotated. These represented both a source of possible training data leakage and labeling noise. To avoid these, we preemptively dropped all lower-cased and stripped duplicates, keeping only the first occurrence. We labeled the kept occurrence with the majority voting of the majority or unanimity labels of the corresponding duplicates.

The official test set comprised $17,558,066$ sentences. Labels were not available for the official test set during the development stage. Indeed, the task's objective was not to classify the relevance of each test sentence for each \ac{BDI} symptom, but instead to retrieve and rank up to $1,000$ sentences from the test set, for each \ac{BDI} symptom. 

\subsection{Fine-Tuning of Foundation Models}

Given the dichotomy in available labels per sentence in the \textit{train} and \textit{val} splits (i.e., the majority and unanimity annotations), we framed foundation model fine-tuning as a regression task to take advantage of all the available data. To that end, we mapped all majority and unanimity labels to a continuous scale between $0-1$ using the mapping function shown in Eq.~\ref{eq:regression-mapping}. This scale encodes the intuition that unanimity labels are closer to the given \ac{BDI} symptom than majority labels. In this regression framework, we fine-tuned the pretrained deberta-v3-large\footnote{\url{https://huggingface.co/microsoft/deberta-v3-large}} \cite{he2021debertav3} foundation model on the \textit{train} split for each of the $21$ \ac{BDI} symptoms, obtaining $21$ fine-tuned models. Each model was fine-tuned for $20$ epochs and the epoch with highest performance on the \textit{val} split was selected. We refer to this as the \textit{mix23} approach. When reverting to the classification setting, outputs $\geq 0.5$ were considered positive (i.e., relevant sentences).


\begin{equation}
    \label{eq:regression-mapping}
    \mathrm{mapping}(x) = 
    \begin{cases}
        0, & \text{if $\mathrm{majority\_label}(x) = 0$} \\
        \frac{2}{3}, & \text{if $\mathrm{majority\_label}(x) = 1$ and $\mathrm{unanimity\_label}(x) = 0$} \\
        1, & \text{if $\mathrm{unanimity\_label}(x) = 1$} \\
    \end{cases}   
\end{equation}

As reflected in Tab.~\ref{tab:annotated-symp}, there were more negative relevance labels than positive. To overcome this, we up-sampled our \textit{train} split by synthesizing positive examples for each \ac{BDI} symptom. Accordingly, for each symptom, we prompted GPT-4o\footnote{\url{https://openai.com/index/hello-gpt-4o}}, Claude Sonnet 3.7\footnote{\url{https://www.anthropic.com/news/claude-3-7-sonnet}}, and Qwen2.5-32B \cite{Qwen2.5} to generate $100$ relevant sentences each. Thus, $300$ positively labeled synthetic sentences were added to the \textit{train} data of each symptom. We prompted these three models, instead of a single one, in order to promote variability in the up-sampling. We performed the same foundation model fine-tuning as described above with the up-sampled data mixed with the original \textit{train} data. We refer to this as the \textit{mix23-aug-1step} approach. We also included another approach, referred to as \textit{mix23-aug-2step}, which further fine-tuned the \textit{mix23-aug-1step} models with just the original data. We performed this second fine-tuning step to ensure that the model observed the original \textit{train} data distribution last.

\subsection{Unsupervised Similarity-Based Approach}

We identified several labeling inconsistencies in the \textit{train} and \textit{val} splits, which motivated us to include an unsupervised approach. Following related work, for each \ac{BDI} symptom, we listed its options and extracted the corresponding embeddings with the all-mpnet-base-v2\footnote{\url{https://huggingface.co/sentence-transformers/all-mpnet-base-v2}} SentenceTransformer. We used the same model to extract all \textit{train} and \textit{val} sentence embeddings. We calculated the maximum similarity between each candidate sentence and the list of options of each \ac{BDI} symptom, obtaining a single cosine-similarity score per sentence in both \textit{train} and \textit{val} splits. 

Contrarily to the related work, we preferred the maximum similarity between a candidate sentence and the list of symptom options, instead of the average similarity, because, assumingly, a sentence does not have to be semantically similar to all symptom options to be considered relevant. This appears especially true given the increase of symptom intensity entailed in the option listing (see examples in Tab.\ref{tab:bdi-example}); indeed, a sentence relevant to the maximum intensity option of a given symptom may very well be semantically distant from the least intense option, the averaging of which would dilute this information, hence our choosing to observe the maximum similarity score. 

We used the similarity scores of \textit{train} sentences for a given \ac{BDI} symptom to define its classification threshold: the average sentence similarity score plus two standard deviations. We mapped all similarity scores to binary labels according to these thresholds. We refer to this as the \textit{maxcos} approach.    

\subsection{Prompt-Based Approaches}

\acp{LLM} have demonstrated impressive zero-shot and few-shot performance in several tasks and domains \cite{Guo2025, gpt4}, which motivated us to explore these approaches for this task. We prompted GPT-4o-Mini whether a given sentence was relevant or not for a given \ac{BDI} symptom. We performed a prompt experimentation stage to arrive at the most adequate prompt wording, but also to gauge performance w.r.t. providing sentence context, $k$-shot prompting ($k \in [0,1,3,5]$), random examples, and semantic similarity examples. We observed the following general behaviors:
\begin{itemize}
    \item Adding sentence \texttt{<PRE>} and \texttt{<POST>} context decreased performance when compared to no context.
    \item With few-shot prompting ($k > 0$):
    \begin{itemize}
        \item Selecting $k$ random examples decreased performance below $0$-shot prompting.
        \item The relevance of the selected examples to the sentence under assessment was crucial for improved performance, i.e., the definition of the semantic similarity strategy.
    \end{itemize}
\end{itemize}

Given these observations, we arrived at a $k$-shot prompting strategy, where the $k$ examples were selected based on their semantic similarity to the sentence under assessment. The pool of exemplars was restricted to the $0$ and $1$ labels of the mapping shown in Eq.~\ref{eq:regression-mapping}. This ensured a clear separation of the two possible outcomes. Note that $2 \times k$ examples are always selected (i.e., $k$ per relevance label). Our prompt is shown in Tab.~\ref{tab:prompt}. The prompt's preamble was based on the task's previous edition official annotation guidelines \cite{Parapar2024}. We refer to this as the \textit{$k$-shot} approach, $k \in [0,1,3,5]$.

\begin{table}
    \centering
    \scriptsize
    \caption{\ac{LLM} prompt template. The exemplars section was omitted in zero-shot prompting.}
    \label{tab:prompt}
    \begin{tabular}{@{} m{\linewidth} @{}}
    \toprule
Consider the following item in Beck's Depression Inventory (BDI-II):\\
\{symptom number\}. \{symptom name\}\\
\{list of symptom options\}\\
 \\
The task consists of annotating sentences in the collection that are topically relevant to the item (relevant to the question and/or to the answers). Note: A relevant sentence should provide some information about the state of the individual related to the topic of the BDI item. But it is not necessary that the exact same words are used. Your job is to assess sentences on how topically relevant they are for the BDI item.\\
 \\
The relevance grades are:\\
1. Relevant: A relevant sentence should be topically related to the BDI item (regardless of the wording) and, additionally, it should refer to the state of the writer about the BDI item.\\
0. Non-Relevant: A non-relevant sentence does not address any topic related to the question and/or the answers of the BDI item (or it is related to the topic but does not represent the writer's state about the BDI item).\\
\\
Together with each sentence, you will receive a set of examples to help with the classification. Answer with just the grade. Use the format [GRADE].\\
\\
Example: \{example sentence\}. Classification: \{example classification\}\\
(... \textit{other examples} ...)\\
\\
Sentence: \{sentence to assess\}. Classification:\\
\bottomrule
    \end{tabular}
\end{table}

\subsection{Evaluation}

The official evaluation is based on \ac{IR} metrics, such as \ac{AP}, \ac{R-PREC}, \ac{P@10}, and \ac{NDCG@1000}. We believe that these metrics cannot be locally implemented due to under-specification. Given the binary labels available in the official training set and data imbalances, we evaluated our approaches under classical classification metrics, namely $F_1$. We designed our approaches to maximize $F_1$ in the \textit{val} split.  

\section{Results and Discussion}
\label{sec:results}

In this section we present and discuss the results of our approaches w.r.t. the development stage, i.e., based on standard classification metrics, namely $F_1$, followed by the official evaluation results of our submissions, which was based on standard \ac{IR} metrics.

\subsection{Development Stage}

Tab.~\ref{tab:results-dev} shows the average $F_1$ performance of the previously described approaches in our development stage (i.e., on the \textit{val} split, described in Tab.~\ref{tab:annotated-symp}). We emphasize again that we framed our development as a classification task, in light of the officially available training annotator majority and unanimity binary labels. Indeed, similar to past edition's official evaluation, we observe model performance in both majority and unanimity annotation settings. The average performance ($\pm$ standard deviation) is across all $21$ \ac{BDI} symptoms. Foundation model fine-tuning approaches (\textit{mix23}, \textit{mix23-aug-1step}, and \textit{mix23-aug-2step}) were the best performing across the board ($\uparrow F_1$), and the most stable across the various symptoms ($\downarrow$ standard deviation). Although there was a small average improvement with \textit{train} data up-sampling, it does not seem to have been critical, as evidenced by the small deltas between \textit{mix23} and \textit{mix23-aug-1step}, and \textit{mix23} and \textit{mix23-aug-2step}. The unsupervised, similarity-based approach (\textit{maxcos}) was the worst performing, with the largest variation across symptoms. We note that zero-shot prompting (\textit{0-shot}) is also unsupervised and the worst performing of the prompt-based methods, although with significantly higher performance than \textit{maxcos}, revealing the positive impact of model size for language representation and encoding (estimated parameter size of GPT-4 $\gg$ all-mpnet-base-v2). The performance of the few-shot prompting approaches (\textit{k-shot}, $k \geq 1$) is aligned with our preliminary findings: performance increases with the number of in-context examples $k$; however, performance does seem to plateau for $k \geq 5$. 

We note that the performance of all approaches dropped significantly from the majority annotation setting to the unanimity one (see the $\Delta$ column in Tab.~\ref{tab:results-dev}). We believe there are two main reasons for this: 1) there were significantly less positively labeled sentences in the unanimity setting, further exacerbating an already unbalanced scenario, and 2) counter-intuitively, we observed in a preliminary stage that the unanimity labels were the most noisy, leading to labeling inconsistencies learned by the models. Regarding these, we see that there was a smaller delta between majority and unanimity performance in \textit{mix23-aug-1step} and \textit{mix23-aug-2step}, than in \textit{mix23}. Under this assessment, it becomes clear that the up-sampling strategy with synthesized examples was critical in improving prediction robustness. The same conclusion can be extrapolated to the prompting strategies, since the delta between majority and unanimity settings decreased as the number of $k$ examples increased. The unsupervised \textit{maxcos} approach had the smallest delta.

\begin{table}
    \centering
    \caption{Development $F_1$ performance of our approaches in both majority and unanimity annotations. \textbf{Bold} (\underline{underline}) values indicate the \textbf{best} (\underline{second best}) performance in majority and unanimity settings. These were obtained by averaging the individual performance for each \ac{BDI} symptom $\pm$ standard deviation.}
    \label{tab:results-dev}
    \begin{tabular}{@{} lccc @{}}
    \toprule
Approach & Majority & Unanimity & $\Delta$ \\
    \midrule
\textit{mix23} & $\mathbf{0.886_{\pm 0.036}}$ & $0.791_{\pm 0.060}$ & $-0.095$ \\
\textit{mix23-aug-1step} & $\underline{0.879_{\pm 0.036}}$ & $\underline{0.804_{\pm 0.055}}$ & $-0.075$ \\
\textit{mix23-aug-2step} & $\mathbf{0.886_{\pm 0.035}}$ & $\mathbf{0.805_{\pm 0.052}}$ & $-0.081$ \\
\textit{maxcos} & $0.775_{\pm 0.052}$ & $0.715_{\pm 0.065}$ & $-0.060$ \\
\textit{0-shot} & $0.822_{\pm 0.053}$ & $0.727_{\pm 0.059}$ & $-0.095$ \\
\textit{1-shot} & $0.830_{\pm 0.046}$ & $0.743_{\pm 0.061}$ & $-0.087$ \\
\textit{3-shot} & $0.842_{\pm 0.048}$ & $0.771_{\pm 0.059}$ & $-0.072$ \\
\textit{5-shot} & $0.846_{\pm 0.050}$ & $0.776_{\pm 0.060}$ & $-0.070$ \\
\bottomrule
    \end{tabular}
\end{table}

\begin{figure}
    \centering
    \includegraphics[width=0.8\linewidth]{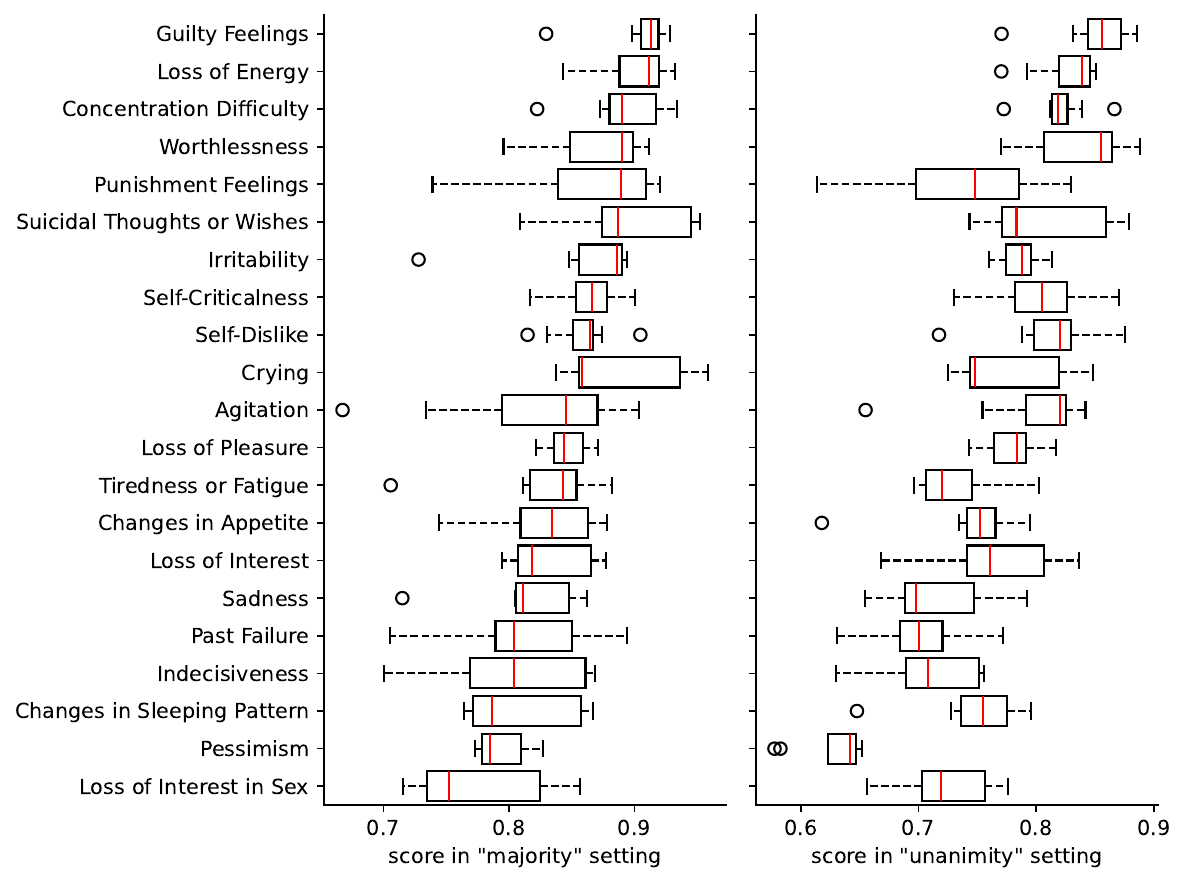}
    \caption{Distribution of \ac{BDI} symptom $F_1$ performance across the multiple approaches, for both majority and unanimity labeling settings. Sorted by descending median value in the majority setting.}
    \label{fig:results-dev-symptoms}
\end{figure}

Fig.~\ref{fig:results-dev-symptoms} shows the $F_1$ performance distribution of each approach, for each \ac{BDI} symptom. Performance varied significantly between symptoms. We emphasize two main measures in this plot: the median value and the distribution wideness. The higher the median value, the better $F_1$ performance for that symptom across methodological approaches. The wider the distribution, the more variation in $F_1$ performance for that symptom across the same approaches. Thus, tight distributions with high median performance are indicative of symptoms that are, overall, ``easy'' to detect (under eRisk's task definition). This includes, e.g., the symptom of Guilty Feelings. Conversely, wide distributions with low median performance are indicative of symptoms that are, overall, ``hard'' to detect. This includes, e.g., the symptoms of Past Failure, Indecisiveness, and Loss of Interest in Sex. We also observe that there were significant distribution differences in certain symptoms, when comparing between the annotation majority and unanimity settings. Symptoms of Agitation, Changes in Sleeping Pattern, and Pessimism are such examples. However, the overall trend (as given by the decreasing order of median performance) was maintained between evaluation settings, suggesting that \ac{BDI} symptoms were equally easy or difficult to detect under both settings. Tab.~\ref{tab:results-best-symptom} complements these results by showing the best performing approach (and corresponding $F_1$ score) for each \ac{BDI} symptom, under both majority and unanimity evaluation settings. This shows that there was not a single best methodological approach for the detection of all \ac{BDI} symptoms. However, as already suggested in Tab.~\ref{tab:results-dev}, the foundation model fine-tuning approaches were by far the most frequently best performing across symptoms. There was only one symptom for which the best performing approach was unsupervised (Loss of Pleasure; \textit{0-shot}).

\begin{table}
    \centering
    \caption{Best performing methodological approach per \ac{BDI} symptom. Sorted by descending best $F_1$ performance in the majority evaluation setting.}
    \label{tab:results-best-symptom}
    \begin{tabular}{@{} l lc lc @{}}
         \toprule
         \multirow{2}{*}{Symptom} & \multicolumn{2}{c}{Best in majority} & \multicolumn{2}{c}{Best in unanimity} \\
         \cmidrule(ll){2-3} \cmidrule(ll){4-5}
         & \multicolumn{1}{c}{Approach} & $F_1$ & \multicolumn{1}{c}{Approach} & $F_1$ \\
         \midrule
Crying & \textit{mix23-aug-2step} & $0.959$ & \textit{mix23-aug-2step} & $0.848$ \\
Suicidal Thoughts or Wishes & \textit{mix23} & $0.952$ & \textit{mix23-aug-2step} & $0.879$ \\
Concentration Difficulty & \textit{mix23} & $0.934$ & \textit{mix23-aug-1step} & $0.867$ \\
Loss of Energy & \textit{mix23} & $0.932$ & \textit{mix23} & $0.851$ \\
Guilty Feelings & \textit{mix23-aug-1step} & $0.928$ & \textit{mix23-aug-2step} & $0.885$ \\
Punishment Feelings & \textit{mix23-aug-1step} & $0.921$ & \textit{mix23-aug-1step} & $0.830$ \\
Worthlessness & \textit{5-shot} & $0.912$ & \textit{5-shot} & $0.888$ \\
Self-Dislike & \textit{mix23} & $0.905$ & \textit{mix23-aug-1step} & $0.875$ \\
Agitation & \textit{mix23} & $0.904$ & \textit{mix23} & $0.842$ \\
Self-Criticalness & \textit{mix23} & $0.901$ & \textit{mix23} & $0.870$ \\
Irritability & \textit{5-shot} & $0.894$ & \textit{mix23-aug-1step} & $0.814$ \\
Past Failure & \textit{mix23-aug-2step} & $0.894$ & \textit{mix23-aug-2step} & $0.772$ \\
Tiredness or Fatigue & \textit{mix23-aug-2step} & $0.882$ & \textit{mix23-aug-2step} & $0.802$ \\
Changes in Appetite & \textit{mix23} & $0.878$ & \textit{mix23-aug-1step} & $0.795$ \\
Loss of Interest & \textit{mix23-aug-2step} & $0.877$ & \textit{mix23} & $0.836$ \\
Loss of Pleasure & \textit{0-shot} & $0.871$ & \textit{mix23-aug-1step} & $0.817$ \\
Indecisiveness & \textit{mix23-aug-2step} & $0.869$ & \textit{mix23} & $0.756$ \\
Changes in Sleeping Pattern & \textit{mix23-aug-2step} & $0.867$ & \textit{mix23} & $0.796$ \\
Sadness & \textit{mix23} & $0.862$ & \textit{mix23-aug-1step} & $0.792$ \\
Loss of Interest in Sex & \textit{mix23-aug-2step} & $0.857$ & \textit{5-shot} & $0.776$ \\
Pessimism & \textit{mix23-aug-1step} & $0.827$ & \textit{mix23} & $0.652$ \\
    \midrule
\textbf{Most frequent best approach (count)} & \multicolumn{2}{c}{\textit{mix23} (8)} & \multicolumn{2}{c}{\textit{mix23} / \textit{mix23-aug-1step} (7)} \\
         \bottomrule
    \end{tabular}
\end{table}

\subsection{Official Submission Evaluation}

Each team was allowed to submit five independent runs to eRisk's task. We designed our submissions according to the development stage results discussed above. Note that the output of the foundation model fine-tuning and similarity-based approaches were self-ranked (fine-tuning was framed as a regression task in the $0-1$ scale). This was not true for prompt-based approaches (output was always binary). Due to the size of the official test set, we used the \textit{maxcos} approach to first filter the candidate test sentences to those that would be positively labeled as relevant by this approach. Our submitted runs were based on the remaining test sentences. Our five runs are detailed below:

\begin{itemize}
    \item \textbf{\textit{mix23}}. This submission consisted entirely of the sorted output with the \textit{mix23} approach described above. For each symptom, we selected up to the first $1,000$ sentences that this approach predicted as positive label.
    
    \item \textbf{\textit{aug-best}}. This submission consisted of obtaining the regression scores of each sentence with both \textit{mix23-aug-1step} and \textit{mix23-aug-2step}, described above, and choosing that which performed best in the development stage per symptom (see Tab.~\ref{tab:results-best-symptom}). For each symptom, we selected up to the first $1,000$ sentences that this approach predicted as positive label.
    
    \item \textbf{\textit{maxcos}}. This submission consisted entirely of the sorted output with the \textit{maxcos} approach described above. For each symptom, we selected up to the first $1,000$ sentences that this approach predicted as positive label (i.e., output $>$ symptom-specific threshold).
    
    \item \textbf{\textit{max}}. This submission consisted of ranking the test candidate sentences according to the maximum score per sentence as given by the previous three submissions (\textit{mix23}, \textit{aug-best}, and \textit{maxcos}). This was an ensemble approach leveraging the findings in Tab.~\ref{tab:results-best-symptom}: indeed, some methods may be particularly better (and more confident) than others, in detecting sentence relevancy. By ranking candidate sentences based on the maximum score of three different approaches, this ensemble prioritizes the individual capacity each approach.

    \item \textbf{\textit{unanimity}}. This submission consisted of selecting only those sentences that were predicted as positive label by all of the first three submissions (\textit{mix23}, \textit{aug-best}, and \textit{maxcos}) and, subsequently, also positively predicted with the prompt-based approach with $k = 5$. These sentences were ranked according to the minimum score of those three submissions. This ensemble approach emphasizes precision and is further conservative in its minimum-score ranking.
\end{itemize}

The official evaluation performance of our runs, according to \ac{IR} metrics, is shown in Tab.~\ref{tab:results-official}. The run \textbf{\textit{unanimity}} performed the best for the \ac{AP}, \ac{R-PREC}, and \ac{P@10} metrics. The run \textbf{\textit{max}} was the best performing for the \ac{NDCG@1000} metric. Notably, both of these runs were ensemble methods, highlighting the importance of leveraging different approaches to capture all the relevant information in the candidate sentences. This was in line with our discussion of results in the development stage. The \ac{NDCG@1000} metric, particularly, emphasizes not only correctly predicted sentences as relevant, but also their ranking; indeed, the run \textbf{\textit{max}} placed first those sentences which one of its ensembled methods was highly confident, thus, performing better than the precision-centric and highly conservative \textbf{\textit{unanimity}} run for this metric. We also note that \textbf{\textit{mix23}} outperformed \textbf{\textit{aug-best}} in all metrics, except \ac{P@10}. The \ac{LLM}-synthesized sentences, used to fine-tune the \textit{mix23-aug-1step} and \textit{mix23-aug-2step} approaches, were fairly obvious w.r.t. their symptom relevance. This may have caused the \textbf{\textit{aug-best}} run to accurately perform for ``obvious'' candidate sentences (hence, the superior \ac{P@10} score), at the cost of under-performing for those that were less ``obvious'' (and hence placed further down the list, not captured by \ac{P@10}). The relative performance of our runs was identical in both majority and unanimity annotation settings. We were the best performing team for all evaluation metrics.

\begin{table}
    \centering
    \caption{Official \ac{IR} performance of our runs. \textbf{Bold} values indicate the \textbf{best} performance in majority and unanimity settings, corresponding to the task winning run. These were provided by this year's official task organizers \cite{Parapar2025}.}
    \label{tab:results-official}
    \begin{tabular}{@{} lcccc @{}}
         \toprule
         \multicolumn{1}{c}{Run} & AP & R-PREC & P@10 & NDCG@1000 \\
         \midrule
         \multicolumn{5}{c}{\textit{{\scriptsize annotator majority evaluation}}} \\
         \midrule
         
         \textit{mix23} & $0.312$ & $0.377$ & $0.643$ & $0.616$ \\
         \textit{aug-best} & $0.247$ & $0.324$ & $0.691$ & $0.560$ \\
         \textit{maxcos} & $0.235$ & $0.320$ & $0.757$ & $0.506$ \\
         \textit{max} & $0.350$ & $0.407$ & $0.648$ & $\mathbf{0.653}$ \\
         \textit{unanimity} & $\mathbf{0.354}$ & $\mathbf{0.433}$ & $\mathbf{0.876}$ & $0.575$ \\

         \midrule
         \multicolumn{5}{c}{\textit{{\scriptsize annotator unanimity evaluation}}} \\
         \midrule

         \textit{mix23} & $0.201$ & $0.279$ & $0.371$ & $0.547$ \\
         \textit{aug-best} & $0.167$ & $0.236$ & $0.414$ & $0.515$ \\
         \textit{maxcos} & $0.164$ & $0.273$ & $0.429$ & $0.472$ \\
         \textit{max} & $0.223$ & $0.308$ & $0.386$ & $\mathbf{0.582}$ \\
         \textit{unanimity} & $\mathbf{0.269}$ & $\mathbf{0.383}$ & $\mathbf{0.509}$ & $0.561$ \\
         
         \bottomrule
    \end{tabular}
\end{table}

\section{Conclusions}
\label{sec:conclusion}

The relation between mental health and linguistic expression opens up opportunities for early detection of depression symptoms in online platforms. eRisk's task of sentence ranking for depression symptoms aims to explore these opportunities. In this work, we discussed our approaches to this year's edition of the task. Our methodology was largely aligned with related work and tackled some of the official data's limitations, such as duplicates, labeling inconsistencies, label imbalances, and labeling dichotomy (i.e., the majority and unanimity annotations). We explored multiple techniques, including foundation model fine-tuning in a regression framework (to leverage all data available in the two annotations) with and without additional synthetic data, similarity-based unsupervised methods, and \ac{LLM} few-shot prompting. Our local development evaluation, based on classification metrics, revealed foundation model fine-tuning as the best performing, followed by few-shot prompting with $k = 5$ examples. Unsupervised similarity-based methods were the worst performing. Based on these results, we submitted five runs for official \ac{IR}-metric evaluation, two of which used ensemble methods. These achieved the highest scores, outperforming submissions from $16$ other teams.

\begin{acknowledgments}
This work was supported by Portuguese national funds through FCT, Fundação para a Ciência e a Tecnologia, under project UIDB/50021/2020 (doi:10.54499/UIDB/50021/2020), and by the Portuguese Recovery and Resilience Plan and Next Generation EU European Funds, through project C644865762-00000008 (Accelerat.AI).
\end{acknowledgments}

\section*{Declaration on Generative AI}
The author(s) have not employed any Generative AI tools.
  

\bibliography{references}

\end{document}